%% file: main.tex
\pdfoutput=1
\documentclass[11pt]{article}
\usepackage[hyphens]{url}   
\usepackage[table]{xcolor}		
\usepackage[preprint]{acl}

\renewcommand\arraystretch{1.0526} %

\usepackage{CJKutf8}
\NewDocumentCommand{\jtext}{m}{\begin{CJK}{UTF8}{ipxm}#1\end{CJK}}

\usepackage[nott]{newtxtext}	
\usepackage{newtxmath}
\usepackage{latexsym}

\usepackage{enumitem}
\setenumerate{noitemsep,topsep=0pt,parsep=0pt,partopsep=0pt}

\usepackage{dblfloatfix}
\usepackage{placeins} 
\usepackage{capt-of}
\usepackage{pbox}

\usepackage[T1]{fontenc}
\usepackage[utf8]{inputenc}

\usepackage{microtype}

\usepackage{enumitem} 
\newlist{paradesc}{description}{1}
\setlist[paradesc]{style=unboxed,leftmargin=0pt,parsep=\parskip,listparindent=\parindent}

\usepackage{array} 
\usepackage{siunitx}
\newcolumntype{C}{>{$}c<{$}}

\usepackage{graphicx}
\usepackage{subcaption} 
\usepackage{booktabs}   
\usepackage{multirow} 
\usepackage{calc}       
\usepackage{ifthen}     
\usepackage{relsize}
\usepackage{needspace}  

\usepackage{relsize} 

\defcitealias{ninjal_2016_construction}{NINJAL}
\defcitealias{ninjal_2018_wordlist}{NINJAL}
\defcitealias{jasso_2024_universities}{JASSO}
\defcitealias{sophiauniversity_2024_levels}{Sophia University}
\newcommand{\citealpaliasyear}[1]{\citetalias{#1}, \citeyear{#1}}
\newcommand{\citepaliasyear}[1]{(\citealpaliasyear{#1})}

\usepackage{xspace}
\newcommand{\MLSJ}{MultiLS-Japanese\xspace}

\title{Difficult for Whom? A Study of Japanese Lexical Complexity}

\author{
    Adam Nohejl\textnormal{\textsuperscript{1}} \quad
    Akio Hayakawa\textnormal{\textsuperscript{2}} \quad
    Yusuke Ide\textnormal{\textsuperscript{1}} \quad
    Taro Watanabe\textnormal{\textsuperscript{1}}
    \\
    \\
    \begin{tabular}{cc}
        \textsuperscript{1}Nara Institute of Science and Technology & \textsuperscript{2}Universitat Pompeu Fabra \\
        $\{\texttt{nohejl.adam.mt3},\;\texttt{ide.yusuke.ja6},\;\texttt{taro}\}\texttt{@is.naist.jp}$ & $\texttt{akio.hayakawa@upf.edu}$ \\
    \end{tabular}%
 }

\begin{document}
\maketitle


\begin{abstract}

The tasks of lexical complexity prediction (LCP) and complex word identification (CWI) commonly presuppose that difficult to understand words are shared by the target population. Meanwhile, personalization methods have also been proposed to adapt models to individual needs. We verify that a recent Japanese LCP dataset is representative of its target population by partially replicating the annotation. By another reannotation we show that native Chinese speakers perceive the complexity differently due to Sino-Japanese vocabulary. To explore the possibilities of personalization, we compare competitive baselines trained on the group mean ratings and individual ratings in terms of performance for an individual. We show that the model trained on a group mean performs similarly to an individual model in the CWI task, while achieving good LCP performance for an individual is difficult. We also experiment with adapting a finetuned BERT model, which results only in marginal improvements across all settings.

\end{abstract}

\section{Introduction}

Complex word identification (CWI) is a task of identifying difficult to understand words in text. CWI systems can be used as components of lexical simplification and readability assessment systems. Lexical complexity prediction (LCP) extends CWI by predicting complexity of words on a continuous scale \citep{shardlow-etal-2020-complex}. 

For both tasks, it is necessary to specify for whom we are predicting the complexity. Non-native speakers have very different needs from people with dyslexia \citep{paetzold-specia-2016-understanding} or children \citep{oshika-etal-2024-simplifying}. For non-native speakers, their L1 background \citep{machida_2001_japanese,ide-etal-2023-japanese} or proficiency level \citep{lee-yeung-2018-personalizing} further determines their needs.

A case has recently been made for personalized CWI, which predicts complex words for an individual \citep{lee-yeung-2018-personalizing,gooding-tragut-2022-one}, and similar methods were earlier proposed for personalized reading assistance \citep{ehara_etal_2013_personalized}. While most research has been done on English as a second language, a personalized CWI system for Chinese as a second language has also been proposed \citep{lee_yeung_2018_automatic}. A shared element of the previously proposed systems is a binary classifier based on a small number of features, such as word frequency or a level from a pedagogical word list. This fits the hypothetical scenario of deployment to user devices and training them using very little labeled data.

Meanwhile, models of increasing size have been applied to lexical complexity prediction targeting relatively wide target populations. In a recent multi-lingual shared task \citep{shardlow-etal-2024-bea}, systems based on large language models (GPT-4) or encoder models (BERT) performed well, especially on relatively high-resource languages such as English or Japanese. The systems were, however, evaluated only on the basis of complexity averaged across all annotators.

We will attempt to answer the following questions for the specific case of the Japanese data employed by the shared task \citep{shardlow-etal-2024-extensible,shardlow-etal-2024-bea}, \MLSJ:
\begin{enumerate}
\item Is the data representative of the intended target population?
\item Can complexity predictions for individuals be improved by training personalized models?
\item How does a simple frequency-based model using a suitable corpus compare to the recent computationally intensive models?
\end{enumerate}

\begin{table*}[t]
\footnotesize
\centering
\setlength\tabcolsep{3.8pt}
\begin{tabular}{
    l<{\hspace{-0.2cm}}
    >{\centering\arraybackslash}p{3.7cm}
    >{\centering\arraybackslash}p{3.7cm}
    >{\centering\arraybackslash}p{3.7cm}
    }
\toprule
 & Original Data & Non-CK L1 Replication & Chinese L1 Reannotation
 \\
\midrule
Native languages &
    English~(5), Swedish~(1), Portuguese~\&~English~(1), French~\&~English~(1), Basque~\&~Spanish~(1), French~(1) &
    Czech~(7), English~\&~Czech~(1), Czech~\&~Ukrainian~(1), Slovak~(1) &
    Chinese (9), Chinese~\&~Cantonese (1)
    \\
\addlinespace[.5em]
JLTP level &
    1~(3), N1~(3), N2~(3), 2~(1) &
    N2~(7), N1 (3) &
    N1~(5), N2~(4), 1~(1)
    \\
Studied Japanese at university &
    \phantom{0}7 of 10 & 10 of 10 & \phantom{0}2 of 10 \\
Currently lives in Japan & 10 of 10 & \phantom{0}0 of 10 & 10 of 10 \\
\addlinespace[.5em]
Lived in Japan (total yrs) & 16.7 (8.3) & \phantom{0}0.7 (0.4) & \phantom{0}4.6 (2.5)\\
Reading in Japanese (hrs/week) & \phantom{0}5.7 (7.6) & \phantom{0}2.6 (2.3) & \phantom{0}9.5 (8.7) \\
Age (yrs) & 40.8 (9.1) & 23.6 (2.7) & 28.2 (2.5) \\
Education (total yrs) & 18.4 (3.7) & 17.2 (2.4) & 19.5 (2.9) \\
Non-native languages & \phantom{0}1.7 (0.5) & \phantom{0}3.1 (1.1) & \phantom{0}2.6 (0.8)\\
\bottomrule
\end{tabular}
\caption{Comparison of the annotator groups of the original data, our replication (same conditions), and our reannotation by Chinese L1 speakers. In the last five rows, we report means followed by standard deviations in parentheses.}\label{tab:replication}
\end{table*}

\section{Analysis}

The \MLSJ dataset is designed as an evaluation dataset consisting of 30 trial instances and 570 test instances. Annotation instructions, annotator profiles, and separate complexity data for each annotator were released online as well.\footnote{\url{https://github.com/naist-nlp/multils-japanese}} Each instance of the dataset is a target word in a sentence context, for which lexical complexity values and simpler substitutions are provided. In this study, we ignore the substitutions as well as the context.

Each instance of both trial and test data was rated by the same set of annotators, which allows us to use the individual ratings in a personalized setting.

\subsection{Target Population}

The annotators were holders of Japanese Language Proficiency Test (JLPT) levels N1 or N2 (or their older equivalents 1 and 2). These levels of JLPT are often required by employers and universities \citepaliasyear{jasso_2024_universities} and have been compared to CEFR levels B2 and C1 \citepaliasyear{sophiauniversity_2024_levels}.
The native language of the annotators was purposely not Chinese or Korean (non-CK), as both languages share a large part of their vocabulary with Japanese.
\citet{maekawa_etal_2014_balanced} estimates the proportion of words of Chinese origin\footnote{
    The traditional terminology for Japanese vocabulary distinguishes between \emph{wago}, indigenous Japanese words; \emph{kango} Sino-Japanese words; and \emph{gairaigo}, foreign words from other languages (e.g.\ English). For simplicity we will call them words of Japanese, Chinese, and other origin, respectively.
} in Japanese text as 17\%  to 47\% based on register. \citet{heo_2010_examination} estimates the proportion of words of Chinese origin in Korean text as 66\%.

\begin{figure}[t]
    \centering
    \includegraphics[width=1.0\linewidth]{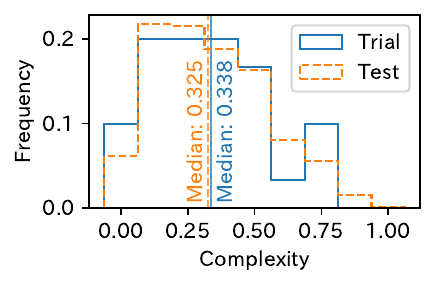}
    \vspace{-2\baselineskip} 
    \caption{Complexity histogram of the trial and test sets.}
    \label{fig:distribution}
\end{figure}

As shown in \autoref{fig:distribution}, the distribution of complexity values in the trial set closely mimics the test set. The distributions of word origins and parts of speech are comparable as well (see \autoref{sec:origin-pos}). We therefore used the trial set to evaluate how representative the dataset is of its target population.
For this purpose we had the trial set reannotated by two groups of annotators: one is from the same target population, while the other has Chinese as their native language. Demographics of each group are summarized in \autoref{tab:replication}.

For the \textbf{non-CK L1 replication} 
we recruited annotators fulfilling the conditions of the original data. Notably, their native languages are neither Chinese nor Korean, but have almost no overlap with native languages of the original annotators. Additionally, while the original annotators have been living in Japan for an average 16.7 years,
for the replication we have recruited undergraduate students or recent graduates of Japanese studies from Charles University in Prague, most of whom have been learning Japanese for 3 to 4 years, out of which no more than 1 year was spent in Japan.

The \textbf{Chinese L1 reannotation} group consists entirely of native Chinese speakers, students or recent graduates of Nara Institute of Science of Technology. The distribution of their proficiency levels is the same as that of the original annotators (six hold JLPT level N1/1 and four hold N2/2). Their mean age and time spent in Japan falls between the means of the original annotators and replication annotators.

\begin{figure}[t]%
    \centering%
    \begin{subfigure}{0.47\linewidth}
        \centering
        \includegraphics[width=0.95\textwidth]{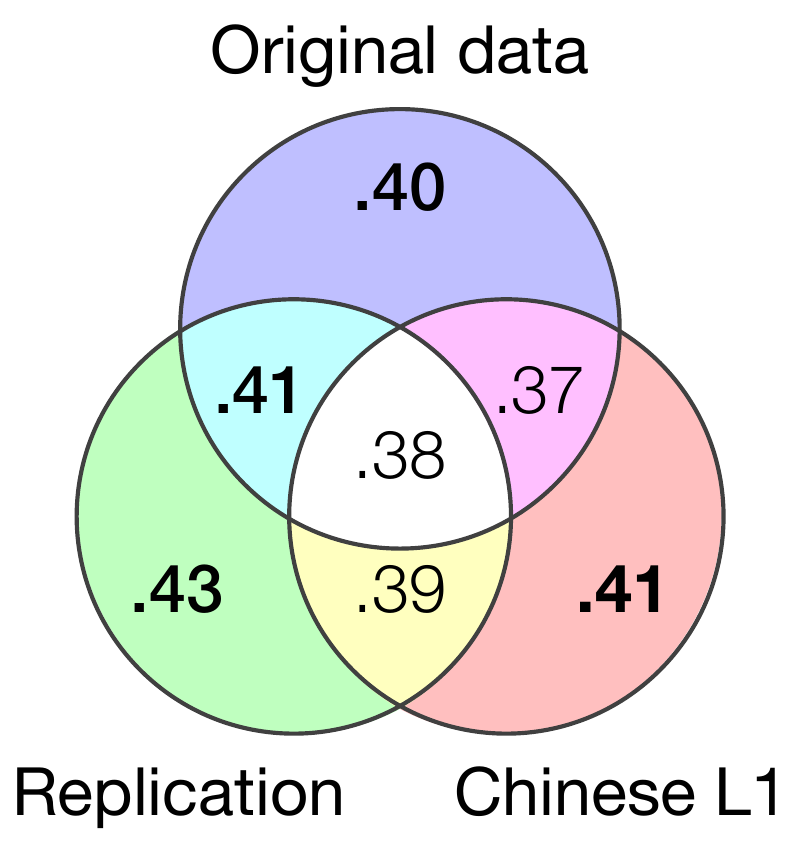}
        \caption{IAA (Krippendorff's $\alpha$)}
        \label{fig:iaa}
    \end{subfigure}%
    \hspace{0.03\linewidth}%
    \begin{subfigure}{0.47\linewidth}
        \centering        \includegraphics[width=0.95\textwidth]{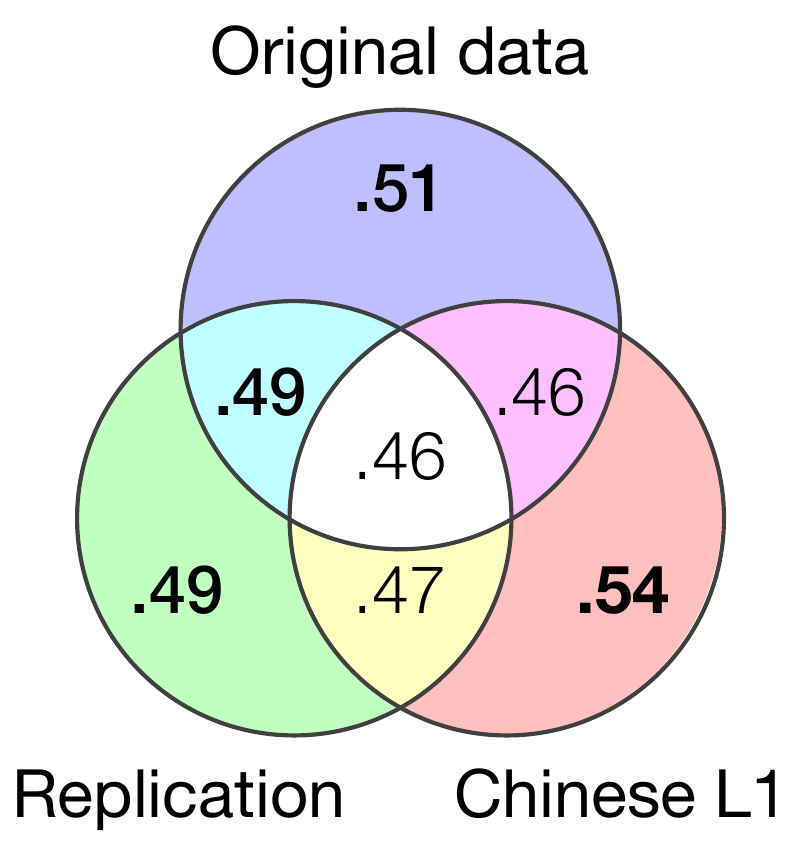}
        \caption{Mean pairwise PCC}
        \label{fig:pcc}
    \end{subfigure}

    \caption{Inter-annotator agreement and mean pairwise correlation in the three annotator groups, the unions of their pairs, and union of all three. Light text denotes that union decreases agreement (correlation).
    }
    \label{fig:iaa-pcc}

\end{figure}

\begin{table*}[t]
\centering
\small
\newlength{\MinusKern}
\setlength{\MinusKern}{\widthof{$+$}-\widthof{$-$}}
\begin{tabular}{lccccc}

\toprule
 &  & TUBELEX & Original Data & \multicolumn{2}{c}{Difference in Complexity From Original Data} \\
\cline{5-6}
Word Origin & \#Words & $\log_{10}$ frequency & Complexity & Non-CK L1 Replication & Chinese L1 Reannotation \\
\midrule
Japanese & 12 & $-5.423$ ($1.427$) & $0.327$ ($0.231$) & $+0.040$ ($0.129$) & $+0.079$ ($0.102$) \\
Chinese &  13 &$-5.247$ ($0.913$)& $0.342$ ($0.180$) & $+0.029$ ($0.119$) & \hspace{\MinusKern}$-0.131$ ($0.093$) \\
\bottomrule
$p$-value & & $0.714$ & $0.866$ & $0.843$ & $\mathbf{< 0.0001}$\\
\end{tabular}
\caption{
    Difference in complexity perceived by two groups of annotators and by the original annotators according to word origin. Mean values are followed by standard deviations in parentheses. We also show log-frequency and original complexity.    
    The $p$-values were obtained from the two-sided unpaired exact permutation test \citep{good_2004_permutation}. Bold font denotes a statistical significant difference in the means between words of Japanese and Chinese origin.
}\label{tab:gap}
\end{table*}

We measured inter-annotator agreement (IAA), using \citeauthor{krippendorff_1970_bivariate}'s \citeyearpar{krippendorff_1970_bivariate} $\alpha$ for interval values, as well as the mean pairwise correlation between annotators. As we can see in  \autoref{fig:iaa-pcc}, the values achieved in both the replication and the Chinese L1 reannotation are similar to the original data. 
When we merge the original data with the replication, however, the inter-annotator agreement does not drop below the agreements in the two groups. In other words, the annotators agree across these two groups as much as within them. This contrasts with the Chinese L1 reannotation, which lowers the agreement when combined with the original data, the replication, or their union. The same applies to the mean pairwise correlation.
A similar tendency for IAA of CK and non-CK L1 annotators was reported by \citet{ide-etal-2023-japanese} for an earlier non-native Japanese LCP dataset, but their native Chinese and Korean annotators also tended to have higher proficiency levels, which complicates the interpretation. In this study, they have the same distribution of proficiency levels.

The underlying cause is a different perception of complexity of words of Japanese and Chinese origin between native Chinese speakers and others. The words perceived as less complex by the native Chinese annotators are almost exclusively words of Chinese origin and vice versa (details provided in \autoref{sec:origin}). As we can see in \autoref{tab:gap}, the gap in complexity perceived by the two groups differs significantly between words of Chinese and Japanese origin. The words of Chinese and Japanese origin do not, however, differ significantly in their frequency, complexity perceived by the original annotators, or the gap between complexity perceived by the original and the Chinese L1 annotators. 

The statistical similarity with annotations by a group with very different demographics supports the hypothesis that the dataset is representative of the target population of non-native Japanese speakers with JLPT proficiency level N2 and higher, whose L1 is not Chinese or Korean.

While the difference between native Chinese speakers and others was to be expected, the similarity with the replication is remarkable. Within the boundaries of target population, we tried to find a homogeneous group of annotators with much less exposure to Japanese language than the original annotators. The students who replicated the trial annotation usually reach level N2 or N1 around their graduation after three to four years of study with limited exposure to Japanese outside their classes. The original annotators not only have lived on average 16.7 years in Japan, but in four cases also acquired their JLPT certificates before year 2010 (as evidenced by old JLPT levels 1 and 2 as opposed to N1 and N2), having ample opportunity to widen their vocabulary beyond the certified level. It is rather surprising how well the two groups agree.

We would like to emphasize that similarly low levels of IAA are common for LCP (e.g.\ $\alpha = 0.32$ or 0.31 reported by \citealp{ide-etal-2023-japanese}), which reflects the subjectivity of the task and shows that there is a room for improvement by personalization.

\subsection{Correlation Analysis}

Word frequency has long been used as a feature for modeling lexical complexity (\citealp{devlin_tait_1998_use}, is an early example). Furthermore, \citet{nohejl_etal_2024_film} demonstrated for multiple languages including Japanese that frequency in TUBELEX, a YouTube subtitle corpus, has a stronger correlation with lexical complexity than frequency in other corpora. We examine correlation with several other variables, not considered by \citet{nohejl_etal_2024_film}. For a fair comparison, we measure Pearson's correlation coefficients (PCC) on the full \MLSJ data. The handling of words missing in data sources is detailed in \autoref{sec:missing}.
We also report ``Potential PCC'' measured only on words present in the individual data sources, thus effectively evaluating each data source on a different subsets of \MLSJ.

As shown in \autoref{tab:corr}, TUBELEX achieves the strongest correlation, followed by the subset of the learner corpus Lang-8 \citep{mizumoto-etal-2011-mining}, where the learners' L1 is not Chinese or Korean (Lang-8-non-CK). The difference between the two is not statistically significant, whereas the difference between Lang-8-non-CK and Lang-8-CK (L1 is Chinese or Korean) is significant.\footnote{Based on \citeauthor{steiger_1980_tests}'s (\citeyear{steiger_1980_tests}) test for dependent correlations with significance level $\alpha = 0.01$.}

Among word frequencies, the weakest correlations are achieved by the corpora CSJ \citepaliasyear{ninjal_2016_construction} and BCCWJ \citep{maekawa_etal_2014_balanced}.
Character frequencies further underperform word frequencies. The Japanese Educational Vocabulary (JEV)\footnote{\url{http://jhlee.sakura.ne.jp/JEV/}} by \citet{sunakawa_etal_2012_construction}, targeting L2 learners, and the WLSP-Familiarity database \citep{asahara-2019-word}, rated by native speakers, have strong potential correlations, but their practical usefulness for LCP is limited by their low coverage, reflected by low actual PCC.

\begin{table}[t]
\small
\centering
\setlength\tabcolsep{3.5pt}
\begin{tabular}{llCC}
\toprule
Variable & Data Source & \text{PCC} & \text{Potential PCC}\\
\midrule
\multirow[l]{5}{*}{\pbox{3cm}{Word\\Log-Frequency}} & {TUBELEX} &\mathbf{-0.66}&\mathbf{-0.66}\\
    & Lang-8-non-CK & -0.64 & -0.64\\
    & Lang-8-CK & -0.61 & -0.61\\
    & CSJ & -0.57 & -0.56\\
    & BCCWJ & -0.55 & -0.57\\
\midrule
L2 Level & JEV &  \phantom{-}0.43 & \phantom{-}0.63\\
\midrule
\multirow[l]{3}{*}{\pbox{3cm}{Character\\Log-Frequency}} & BCCWJ &-0.35 & -0.37\\
    & Lang-8-non-CK & -0.35 &-0.36\\
    & Lang-8-CK & -0.33 &-0.34\\
\midrule
L1 Familiarity & WLSP (Asahara) &  -0.23  & -0.55\\
\bottomrule
\end{tabular}
\caption{Correlation (PCC) of \MLSJ test set complexity with log-frequencies, learner levels and native familiarity. For BCCWJ, CSJ, JEV, and WLSP, values were looked up by lemma. Potential PCC only considers words present in each data source. 
Rows are ordered by PCC strength (absolute value): naturally, \emph{high} complexity is associated with \emph{low} frequency and familiarity, hence the negative values.
}
\label{tab:corr}
\end{table}

\section{Experiments}
\label{sec:experiments}

Following the design of \MLSJ, we use the 30 trial instances for training, and the 570 test instances for evaluation. We only use the datasets original data, not the replication or reannotation, for the experiments. 

We evaluate models in four settings determined by training and test data, e.g.\ the ``Group-Individual'' denotes training on group data (mean for LCP or majority class for CWI) and evaluation on individual data. With the exception of the Group-Group setting, where a single model is evaluated on a single test set, we therefore report the results as means and standard deviations. In the case of Individual-Individual, we evaluate each model trained on individual data only on the corresponding individual test data.

We also evaluate models in the CWI task by considering complexity values $\ge 0.375$ (the midpoint between the \emph{easy} and \emph{neutral} ratings in \MLSJ) to be complex. The results in CWI are easier to interpret, and can be compared with previous personalized CWI research. 
In addition to CWI models (binary classifiers), we also evaluate LCP models in CWI (henceforth LCP-CWI) by interpreting their values as the positive class if they exceed the threshold. 

For LCP, we measure $R^2$, the coefficient of determination. For CWI, we measure performance using macro-averaged F1 score, i.e.\ the average of F1 scores for the positive and negative class, in line with previous research \citep{yimam-etal-2018-report,gooding-tragut-2022-one}.

Detailed information about the experimental models is provided in \autoref{sec:models}.

\subsection{Frequency Baseline}

As a baseline for LCP, we fit a linear regression using log-frequency in TUBELEX to the trial data.
As shown in \autoref{tab:tubelex-r2}, the model performs well in the Group-Group setting (0.41), on par with the best $R^2$ result for Japanese in the shared task (0.413) obtained using a GPT-4-based model \citep{enomoto-etal-2024-tmu}.

If we, however, train and evaluate the same baseline on individual data, the performance drops drastically (0.13).
This may be counter-intuitive, as we are training and evaluating on the data annotated by the same individual, but it shows that that the strong correlation with log-frequency, and consequently the good performance of the baseline on group data, is mostly a result of individual idiosyncrasies being smoothed out by the group average. For LCP, the personalized Individual-Individual frequency baseline did not fare well. Results in the other settings were even worse with mean $R^2$ below zero.

\begin{table}[t]
\small
\centering
\include{r-tubelex-r2}
\caption{LCP results ($R^2$) using TUBELEX log-frequency as a single feature.}
\label{tab:tubelex-r2}
\end{table}

\begin{table}[t]
\small
\centering
\include{r-tubelex-mf1}
\caption{CWI results (F1) using TUBELEX log-frequency as a single feature.}
\label{tab:tubelex-mf1}
\end{table}

For CWI, we fit a logistic regression model using the same single feature, and compare it with the LCP model, evaluated as LCP-CWI.
As in the previous case, the results in \autoref{tab:tubelex-mf1} show that both kinds of models perform worse in the Individual-Individual setting than in the Group-Group setting, although the difference is smaller in CWI. Surprisingly, however, the CWI model in the Group-Individual setting reaches almost the same F1 score as personalized Individual-Individual CWI models. 
Additionally, the LCP model in the Group-Individual setting is very competitive when evaluated as LCP-CWI (0.65), outperforming the personalized LCP model (0.56) and nearing the performance of personalized CWI models (0.67). While it is difficult to predict the exact complexity in LCP, models trained on the group perform relatively well in the CWI task, even for individuals. 

\subsection{BERT-Based Model}


The target population of \MLSJ is similar to that of non-CK L1 data of the Japanese Lexical Complexity for Non-Native Readers (JaLeCoN) dataset \citep{ide-etal-2023-japanese}. 
We finetuned the BERT model described by \citeauthor{ide-etal-2023-japanese} for CK and non-CK data of the whole JaLeCoN dataset. 
To adapt it to \MLSJ, we used its output (predicted complexity) as a feature for linear and logistic regression either alone or together with the TUBELEX log-frequency. \autoref{sec:appendix} provides results of all variants.

The best results, shown in Tables~\ref{tab:tubelex-jnc-r2} and  \ref{tab:tubelex-jnc-mf1}, were achieved by combining frequency with the model finetuned on JaLeCon-non-CK. All settings achieved only a marginal improvement over the frequency baseline. 

\begin{table}[t]
\small
\centering
\include{r-tubelex-jnc-r2}
\caption{LCP results ($R^2$) using TUBELEX log-frequency and output of the BERT model trained on JaLeCoN-non-CK.}
\label{tab:tubelex-jnc-r2}
\end{table}

\begin{table}[t]
\small
\centering
\include{r-tubelex-jnc-mf1}
\caption{CWI results (F1) using TUBELEX log-frequency and output of the BERT model trained on JaLeCoN-non-CK.}
\label{tab:tubelex-jnc-mf1}
\end{table}

\section{Conclusion}

We demonstrated that the \MLSJ dataset is representative of its intended target population by comparing its IAA and correlation  with an annotation replicated by a group with different demographics but fulfilling the conditions of proficiency and not having a Chinese or Korean L1 background.

Additionally, we demonstrated a clear difference in complexity perception of Japanese words, based on word origin, between this population and native Chinese speakers of the same proficiency levels in Japanese. To which extent this applies to native Korean speakers is a question for future research.

We found that achieving good performance in individual LCP is more difficult than in individual CWI. In individual LCP, personalization resulted in a small improvement over training on group data, but in individual CWI, personalization and training on group data performed similarly well.

The TUBELEX frequency baseline performed on par with the \mbox{GPT-4}-based model that achieved the best result in a recent shared task. Combining the frequency feature with a fine-tuned BERT model resulted only in marginal improvements in  both the group and the individual setting.

In future work, we would like to investigate the effect of larger training data paired with additional features (e.g.\ register of a word) and the performance of different methods of sampling training data, such as uncertainty sampling.

\section*{Lay Summary}

To make text easier to understand using an automated system, it is necessary to identify difficult words, which depends on the text's reader. The automated systems, therefore, need to focus on a specific target population, such as non-native speakers, or be personalized for the reader. The difficulty of words is called ``lexical complexity'' and can be rated on a scale.

The performance of systems for estimating lexical complexity can be scored using specialized datasets in which complexity is rated by people from the target population. The systems for estimating lexical complexity are usually scored based on the average rating by a group from the target population, not individuals. Additionally, some of the best performing systems use large language models such as GPT-4, which are costly to run.

Our study uses a dataset targeting highly proficient non-native Japanese speakers, excluding native Chinese and Korean speakers, who would have the advantage of knowing vocabulary shared among the three languages. We explore the following questions:

\begin{enumerate}
\item Is the dataset representative of the target population?
\item Can personalized systems improve estimates over those for the group average?
\item How does a simple word-frequency-based system compare to the costlier models?
\end{enumerate}

By having the data rerated by two new groups, we confirmed that the dataset represents the target group well and that native Chinese speakers perceive Japanese complexity differently.

We compared personalized systems and systems based on the group average in terms of performance for individuals in two scenarios: When estimating lexical complexity rated on a scale, personalized systems performed slightly better. When we only classified the words as difficult or not difficult, the systems based on the group average and the personalized ones performed similarly. Regardless of the system or the scenario, we found it much more challenging to achieve good performance for the individuals than for the group average, which smooths out individual idiosyncrasies.

A simple frequency-based system using word frequency in YouTube subtitles slightly outperformed a recent model based on GPT-4, which is much more expensive to run.

In future work, we would like to investigate the effect of larger training data paired with more complex systems, which would consider other features of the words, such as register (formal vs. informal).

\section*{Limitations}

We focused on a specific target population of non-native speakers defined by the exclusion of two specific L1s and relatively high proficiency levels. Even the simple personalization methods, which did not perform particularly well in our setting, may provide an advantage for a more diverse population, effectively providing adaptation to large differences in proficiency.
We also have not evaluated different methods of training data sampling (e.g.\ uncertainty sampling in an active learning scenario, which may improve performance while using the same size of training data). We only performed objective metric-based evaluation of the system's performance. An additional human evaluation would also be desirable.

\section*{Acknowledgments}

The \MLSJ dataset was created as a part of a joint research project of Nara Institute of Science and Technology and Nikkei, Inc. We would like to express our sincere gratitude to the volunteer annotators for their cooperation on partial replication and reannotation. The replication was performed by students and graduates of Japanese Studies at Charles University. The reannotation by native Chinese speakers was performed by students and graduates of Nara Institute of Science and Technology. We would like to thank Petra Kanasugi and Huayang Li for helping us recruit the annotators. We are grateful to the anonymous reviewers for their insightful comments and helpful suggestions.

\bibliography{anthology,ja_ls_lcp_custom}

\clearpage
\appendix
\onecolumn
\section{Comparison of Word Origins and Parts of Speech in the Test and Trial Sets}
\label{sec:origin-pos}

\begin{table}[h]
\small
\centering
\include{origin-pos}
\caption{
Comparison of the test set and trial set in terms of proportions of words containing tokens of Chinese or English origin and parts of speech. The remaining target words are purely of indigenous Japanese origin. We distinguish between adjectives (\jtext{形容詞}, so-called \emph{i}-adjectives) and adjectival nouns (\jtext{形容動詞} or \jtext{形状詞}, \emph{na}-adjectives and \emph{to}/\emph{taru}-adjectives). The Particle category excludes conjunctive particles (\jtext{接続助詞}), which we categorize as Auxiliaries together with auxiliary verbs (\jtext{助動詞}). MWE are multi-word expressions, typically noun-verb phrases.
}
\label{tab:origin-pos}
\end{table}

\vfill

\section{Handling of Words Missing in Data Sources}
\label{sec:missing}

\begin{table}[h]
\small
\setlength\tabcolsep{4pt}
\centering
\newlength{\FormulaW}
\setlength{\FormulaW}{4.9cm}
\begin{tabular}{lllp{\FormulaW}l}
\toprule
 &  & {Handling of} & & Sequence of Tokens \\

{Data Source} & {Values} & {Missing Values} & {Formula for One Token or Character $x$} & or Characters $\mathbf{s}$ \\
\midrule

All Corpora & Log-Frequency & Laplace smoothing &
\parbox{\FormulaW}{$\displaystyle 
        f(x) = \log \left( \frac{\textrm{count}(x) + 1}{\textrm{\#tokens} + \textrm{\#types} } \right)
    $} &
    $\displaystyle
    f(\mathbf{s}) = \min_{x \in \mathbf{s}} f(x)
    $
    \\

\specialrule{\arrayrulewidth}{2\aboverulesep}{2\belowrulesep}
    
JEV & Levels 1--6 & Dummy values & \parbox{\FormulaW}{$\displaystyle
f(x) = \begin{cases}
    \textrm{level}(x) & \textrm{if } x \in \textrm{JEV}\\
    7 & \textrm{otherwise}\\
\end{cases}
$
}&
    $\displaystyle
    f(\mathbf{s}) = \max_{x \in \mathbf{s}} f(x)
    $
    \\

\specialrule{\arrayrulewidth}{2\aboverulesep}{2\belowrulesep}

WLSP-Familiarity & $F \subset \mathbb{R}$  & Dummy values & \parbox{\FormulaW}{$\displaystyle
f(x) = \begin{cases}
    \textrm{familiarity}(x) & \textrm{if } x \in \textrm{WLSP}\\
    \textrm{min}(F) & \textrm{otherwise}\\
\end{cases}
$
}&
    $\displaystyle
    f(\mathbf{s}) = \min_{x \in \mathbf{s}} f(x)
    $
    \\
\specialrule{0pt}{\aboverulesep}{0pt}

\bottomrule
\end{tabular}

\caption{Handling of words (or characters) missing in data sources used for PCC computation in \autoref{tab:corr}. For all corpora, we use Laplace smoothing recommended by \citet{brysbaert_diependaele_2013_dealing} to provide log-frequency values even for words missing in the corpora. To words missing in JEV, we assign the value corresponding to a level beyond those present in the data. To words missing in WLSP-Familiarity, we assign the minimum familiarity level present in the data. To sequences consisting of multiple tokens or characters, we assign the minimum or maximum value assigned to the individual items as appropriate.
}
\end{table}
\vfill

\clearpage
\newcommand{\OC}{\color[HTML]{FF0000}{Chinese}}
\newcommand{\OCJ}{\color[HTML]{FF00FF}{Ch.\ + Ja.}}
\newcommand{\OJ}{\color[HTML]{0000FF}{Japanese}}
\newcommand{\OEn}{\color[HTML]{00FF00}{English}}

\section{Difference of Complexity Perception by Annotators' L1 and Word Origin}
\label{sec:origin}
\subsection{Original Annotation and Chinese L1 Reannnotation}
\renewcommand\arraystretch{1.049}

\begin{table*}[h]
\centering\small
\input{diff-chinese-l1}
\caption{Target words in the trial set of \MLSJ; their word origin; log-frequency; mean complexity annotated by the original annotators, whose L1 was neither Chinese or Korean, and the Chinese L1 annotators;  difference between the former and the latter. The table is sorted by the complexity difference to highlight the overlap between words of Chinese origin and words perceived as less complex by the Chinese L1 annotators compared to the original annotators. ``Ch.\ + Ja.'' denotes expressions mixing content words of Chinese and Japanese origin. We ignore the origin of common functional words such as particles and light verbs.
}
\label{tab:diff-chinese-l1}
\end{table*}

\vfill
\noindent%
\begin{minipage}{\textwidth}%
\captionsetup{type=table}%
    \setlength\tabcolsep{4pt}
    \footnotesize
    \centering
    \includegraphics[width=0.84\linewidth]{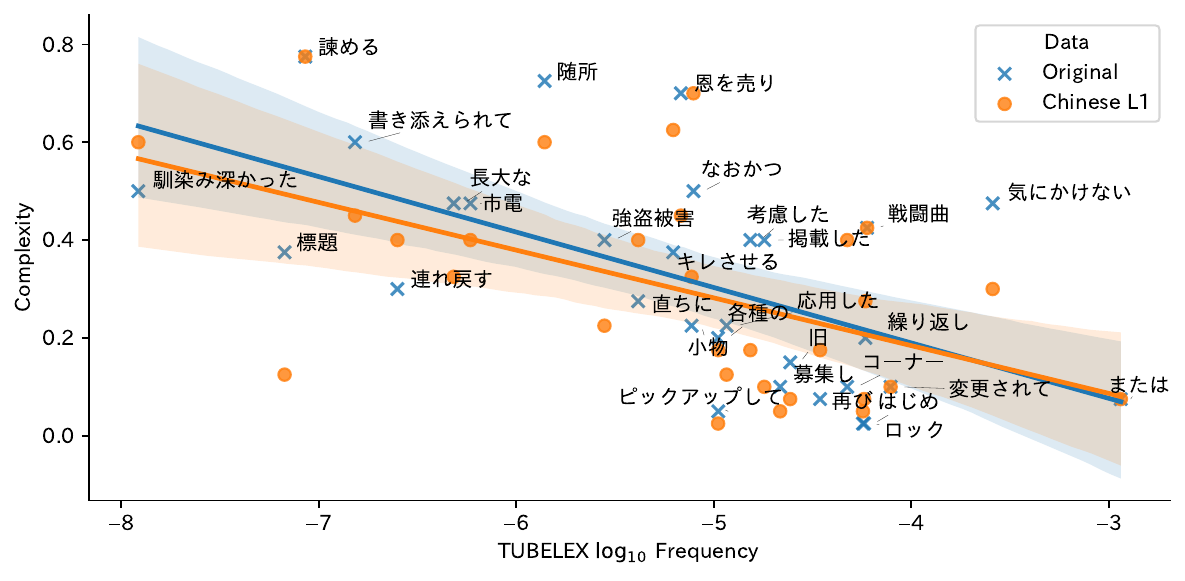}
    \vspace{-0.75\baselineskip}
    \captionof{figure}{Mean complexity of target words in the the trial set of \MLSJ and in the Chinese L1 reannotation, plotted against log-frequency. Lines show linear fit with 95\% confidence interval as a shaded area.}
    \label{fig:freq-comp-zh}
\end{minipage}

\clearpage 
\subsection{Original Annotation and Replication}

\begin{table*}[h]
\centering\small
\input{diff-replication}
\caption{Target words in the trial set of \MLSJ; their word origin; log-frequency; mean complexity annotated by the original annotators, and the replication annotators; difference between the mean complexities perceived by the two groups. Neither annotator group contained native Chinese or Korean speakers, hence compared to \autoref{tab:diff-chinese-l1}, there is not any clear tendency for words of Chinese origin.
}
\end{table*}

\vfill
\noindent%
\begin{minipage}{\textwidth}%
\captionsetup{type=table}%
    \setlength\tabcolsep{4pt}
    \footnotesize
    \centering
    \includegraphics[width=0.84\linewidth]{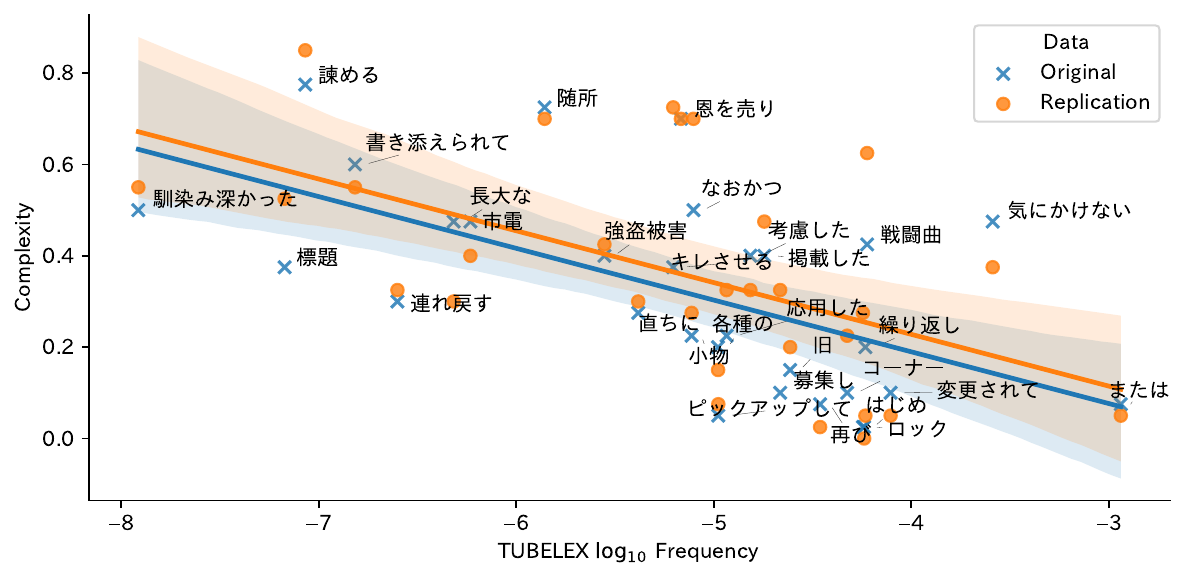}
    \vspace{-0.75\baselineskip}
    \captionof{figure}{Mean complexity of target words in the the trial set of \MLSJ and in the replication. Lines show linear fit with 95\% confidence interval as a shaded area.}
    \label{fig:freq-comp}
\end{minipage}

\clearpage

\section{Model Details}
\renewcommand\arraystretch{1.0526}
\label{sec:models}

\begin{table*}[h]
\centering
\begin{tabular}{lp{0.25\textwidth}p{0.3\textwidth}p{0.25\textwidth}}
\toprule
Task & Model Description & Implementation & Postprocessing\\
\midrule

LCP & Linear regression with L2 regularization ($\alpha = 1$) & \texttt{Ridge()} & Clip values to the valid range [0, 1].\\
\hline
CWI & Logistic regression with balanced class weights & \texttt{LogisticRegression(}\newline
\mbox{\hspace{\widthof{\texttt{-}}}}\texttt{class\_weight='balanced')} &
---\\
\bottomrule
\end{tabular}
\caption{
Details of the models used for experiments in \autoref{sec:experiments}, implemented using the \texttt{scikit-learn} Python package, namely classes from \texttt{sklearn.linear\_model}. For baselines, the only feature is log-frequency in TUBELEX (see \autoref{sec:missing}). For the BERT-based models, the features are (1) output of the finetuned BERT model and optionally (2) log-frequency in TUBELEX. The BERT models are exactly as described by \citet{ide-etal-2023-japanese}, except that we finetuned them for CK and non-CK complexity of the whole JaLeCoN dataset (not using any train-test split of the data).
}
\end{table*}

\vfill

\section{Results of the BERT-based Model Variants}
\label{sec:appendix}

\begin{table}[h]
\centering
\include{r-jnc-r2}
\caption{LCP results ($R^2$) using output of the BERT model trained on JaLeCoN-non-CK.}
\end{table}

\begin{table}[h]
\centering
\include{r-jnc-mf1}
\caption{CWI results (F1) using output of the BERT model trained on JaLeCoN-non-CK.}
\end{table}
\clearpage%
\begin{table}[h]
\centering
\include{r-tubelex-jc-r2}
\caption{LCP results ($R^2$) using TUBELEX log-frequency and output of the BERT model trained on JaLeCoN-CK.}
\end{table}
\begin{table}[h]
\centering
\include{r-tubelex-jc-mf1}
\caption{CWI results (F1) using TUBELEX log-frequency and output of the BERT model trained on JaLeCoN-CK.}
\end{table}
\begin{table}[h]
\centering
\include{r-jc-r2}
\caption{LCP results ($R^2$) using output of the BERT model trained on JaLeCoN-CK.}
\end{table}
\vspace{\baselineskip}
\noindent%
\begin{minipage}{\textwidth}%
\captionsetup{type=table}%
    \setlength\tabcolsep{4pt}
    \footnotesize
    \centering
\include{r-jc-mf1}
\captionof{table}{CWI results (F1) using output of the BERT model trained on JaLeCoN-CK.}
\end{minipage}
\end{document}

%% file: r-tubelex-r2.tex
\begin{tabular}{lS[table-format=-1.2(1.2)]S[table-format=-1.2(1.2)]}
\toprule
{Test} & {Group} & {Individual} \\
{Train} & {} & {} \\
\midrule
Group & \bfseries 0.41 & -0.10(0.41) \\
Individual & -0.06(0.64) & \bfseries 0.13(0.15) \\
\bottomrule
\end{tabular}

%% file: r-tubelex-mf1.tex
\begin{tabular}{llS[table-format=1.2(1.2)]S[table-format=1.2(1.2)]}
\toprule
{} & {Test} & {Group} & {Individual} \\
{Model} & {Train} & {} & {} \\
\midrule
\multirow[c]{2}{*}{LCP-CWI} & Group & 0.71 & 0.65(0.05) \\
 & Individual & 0.56(0.18) & 0.56(0.11) \\
\hline
\multirow[c]{2}{*}{CWI} & Group & \bfseries 0.78 & 0.67(0.06) \\
 & Individual & 0.77(0.02) & \bfseries 0.67(0.04) \\
\bottomrule
\end{tabular}

%% file: r-tubelex-jnc-r2.tex
\begin{tabular}{lS[table-format=-1.2(1.2)]S[table-format=-1.2(1.2)]}
\toprule
{Test} & {Group} & {Individual} \\
{Train} & {} & {} \\
\midrule
Group & \bfseries 0.43 & -0.08(0.41) \\
Individual & -0.04(0.65) & \bfseries 0.15(0.15) \\
\bottomrule
\end{tabular}

%% file: r-tubelex-jnc-mf1.tex
\begin{tabular}{llS[table-format=1.2(1.2)]S[table-format=1.2(1.2)]}
\toprule
{} & {Test} & {Group} & {Individual} \\
{Model} & {Train} & {} & {} \\
\midrule
\multirow[c]{2}{*}{LCP-CWI} & Group & 0.72 & 0.66(0.05) \\
 & Individual & 0.57(0.19) & 0.57(0.12) \\
\hline
\multirow[c]{2}{*}{CWI} & Group & \bfseries 0.79 & \bfseries 0.67(0.06) \\
 & Individual & 0.77(0.02) & 0.67(0.04) \\
\bottomrule
\end{tabular}

%% file: origin-pos.tex
\begin{tabular}{llrr}
\toprule
 &  & Test & Trial \\
\midrule
\multirow[c]{2}{*}{Word Origin} & Chinese & 55.4\% & 50.0\% \\
 & English & 5.3\% & 10.0\% \\
\hline
\multirow[c]{12}{*}{Part of Speech} & Noun & 45.6\% & 36.7\% \\
 & Verb & 27.5\% & 36.7\% \\
 & Adjectival Noun & 7.4\% & 3.3\% \\
 & MWE & 7.0\% & 6.7\% \\
 & Adverb & 6.0\% & 10.0\% \\
 & Adjective & 2.1\% & 3.3\% \\
 & Particle & 1.8\% & --- \\
 & Pronoun & 0.9\% & --- \\
 & Conjunction & 0.7\% & 3.3\% \\
 & Suffix & 0.5\% & --- \\
 & Auxiliary & 0.4\% & --- \\
 & Prefix & 0.2\% & --- \\
\bottomrule
\end{tabular}

%% file: diff-chinese-l1.tex
\begin{tabular}{lcS[table-format=-1.3]S[table-format=1.3]S[table-format=1.3]S[table-format=-1.3,retain-explicit-plus]}
\toprule
{} & {} & {TUBELEX} & \multicolumn{3}{c}{Complexity} \\
\cline{4-6}
{Target Word} & {Word Origin} & {$\log_{10}$ Frequency} & {Original} & {Chinese L1} & {Difference \textcolor{gray}{$\downarrow^{\raisebox{-0.1em}{\small\hspace{0.03em}$-$}}_{\raisebox{0.06em}[0em]{\small\hspace{0.04em}$+$}}$}} \\
\midrule
\jtext{掲載した} & \OC & -4.744 & 0.400 & 0.100 & -0.300 \\
\jtext{恩を売り} & \OCJ & -5.166 & 0.700 & 0.450 & -0.250 \\
\jtext{標題} & \OC & -7.173 & 0.375 & 0.125 & -0.250 \\
\jtext{考慮した} & \OC & -4.815 & 0.400 & 0.175 & -0.225 \\
\jtext{強盗被害} & \OC & -5.554 & 0.400 & 0.225 & -0.175 \\
\jtext{各種の} & \OC & -4.978 & 0.200 & 0.025 & -0.175 \\
\jtext{気にかけない} & \OCJ & -3.588 & 0.475 & 0.300 & -0.175 \\
\jtext{書き添えられて} & \OJ & -6.817 & 0.600 & 0.450 & -0.150 \\
\jtext{長大な} & \OC & -6.317 & 0.475 & 0.325 & -0.150 \\
\jtext{随所} & \OC & -5.857 & 0.725 & 0.600 & -0.125 \\
\jtext{応用した} & \OC & -4.935 & 0.225 & 0.125 & -0.100 \\
\jtext{旧} & \OC & -4.613 & 0.150 & 0.075 & -0.075 \\
\jtext{市電} & \OC & -6.232 & 0.475 & 0.400 & -0.075 \\
\jtext{募集し} & \OC & -4.664 & 0.100 & 0.050 & -0.050 \\
\jtext{諫める} & \OJ & -7.068 & 0.775 & 0.775 & 0.000 \\
\jtext{変更されて} & \OC & -4.105 & 0.100 & 0.100 & 0.000 \\
\jtext{または} & \OJ & -2.939 & 0.075 & 0.075 & 0.000 \\
\jtext{戦闘曲} & \OC & -4.224 & 0.425 & 0.425 & 0.000 \\
\jtext{ロック} & \OEn & -4.245 & 0.025 & 0.050 & +0.025 \\
\jtext{はじめ} & \OJ & -4.239 & 0.025 & 0.075 & +0.050 \\
\jtext{繰り返し} & \OJ & -4.232 & 0.200 & 0.275 & +0.075 \\
\jtext{小物} & \OJ & -5.112 & 0.225 & 0.325 & +0.100 \\
\jtext{馴染み深かった} & \OJ & -7.913 & 0.500 & 0.600 & +0.100 \\
\jtext{再び} & \OJ & -4.462 & 0.075 & 0.175 & +0.100 \\
\jtext{連れ戻す} & \OJ & -6.602 & 0.300 & 0.400 & +0.100 \\
\jtext{ピックアップして} & \OEn & -4.977 & 0.050 & 0.175 & +0.125 \\
\jtext{直ちに} & \OJ & -5.383 & 0.275 & 0.400 & +0.125 \\
\jtext{なおかつ} & \OJ & -5.103 & 0.500 & 0.700 & +0.200 \\
\jtext{キレさせる} & \OJ & -5.205 & 0.375 & 0.625 & +0.250 \\
\jtext{コーナー} & \OEn & -4.325 & 0.100 & 0.400 & +0.300 \\
\bottomrule
\end{tabular}

%% file: diff-replication.tex
\begin{tabular}{lcS[table-format=-1.3]S[table-format=1.3]S[table-format=1.3]S[table-format=-1.3,retain-explicit-plus]}
\toprule
{} & {} & {TUBELEX} & \multicolumn{3}{c}{Complexity} \\
\cline{4-6}
{Target Word} & {Word Origin} & {$\log_{10}$ Frequency} & {Original} & {Replication} & {Difference \textcolor{gray}{$\downarrow^{\raisebox{-0.1em}{\small\hspace{0.03em}$-$}}_{\raisebox{0.06em}[0em]{\small\hspace{0.04em}$+$}}$}} \\
\midrule
\jtext{長大な} & \OC & -6.317 & 0.475 & 0.300 & -0.175 \\
\jtext{繰り返し} & \OJ & -4.232 & 0.200 & 0.050 & -0.150 \\
\jtext{気にかけない} & \OCJ & -3.588 & 0.475 & 0.375 & -0.100 \\
\jtext{考慮した} & \OC & -4.815 & 0.400 & 0.325 & -0.075 \\
\jtext{市電} & \OC & -6.232 & 0.475 & 0.400 & -0.075 \\
\jtext{変更されて} & \OC & -4.105 & 0.100 & 0.050 & -0.050 \\
\jtext{各種の} & \OC & -4.978 & 0.200 & 0.150 & -0.050 \\
\jtext{再び} & \OJ & -4.462 & 0.075 & 0.025 & -0.050 \\
\jtext{書き添えられて} & \OJ & -6.817 & 0.600 & 0.550 & -0.050 \\
\jtext{随所} & \OC & -5.857 & 0.725 & 0.700 & -0.025 \\
\jtext{または} & \OJ & -2.939 & 0.075 & 0.050 & -0.025 \\
\jtext{はじめ} & \OJ & -4.239 & 0.025 & 0.000 & -0.025 \\
\jtext{恩を売り} & \OCJ & -5.166 & 0.700 & 0.700 & 0.000 \\
\jtext{連れ戻す} & \OJ & -6.602 & 0.300 & 0.325 & +0.025 \\
\jtext{ピックアップして} & \OEn & -4.977 & 0.050 & 0.075 & +0.025 \\
\jtext{強盗被害} & \OC & -5.554 & 0.400 & 0.425 & +0.025 \\
\jtext{直ちに} & \OJ & -5.383 & 0.275 & 0.300 & +0.025 \\
\jtext{小物} & \OJ & -5.112 & 0.225 & 0.275 & +0.050 \\
\jtext{旧} & \OC & -4.613 & 0.150 & 0.200 & +0.050 \\
\jtext{馴染み深かった} & \OJ & -7.913 & 0.500 & 0.550 & +0.050 \\
\jtext{掲載した} & \OC & -4.744 & 0.400 & 0.475 & +0.075 \\
\jtext{諫める} & \OJ & -7.068 & 0.775 & 0.850 & +0.075 \\
\jtext{応用した} & \OC & -4.935 & 0.225 & 0.325 & +0.100 \\
\jtext{コーナー} & \OEn & -4.325 & 0.100 & 0.225 & +0.125 \\
\jtext{標題} & \OC & -7.173 & 0.375 & 0.525 & +0.150 \\
\jtext{なおかつ} & \OJ & -5.103 & 0.500 & 0.700 & +0.200 \\
\jtext{戦闘曲} & \OC & -4.224 & 0.425 & 0.625 & +0.200 \\
\jtext{募集し} & \OC & -4.664 & 0.100 & 0.325 & +0.225 \\
\jtext{ロック} & \OEn & -4.245 & 0.025 & 0.275 & +0.250 \\
\jtext{キレさせる} & \OJ & -5.205 & 0.375 & 0.725 & +0.350 \\
\bottomrule
\end{tabular}

%% file: r-jnc-r2.tex
\begin{tabular}{lS[table-format=-1.2(1.2)]S[table-format=-1.2(1.2)]}
\toprule
{Test} & {Group} & {Individual} \\
{Train} & {} & {} \\
\midrule
Group & \bfseries 0.14 & -0.24(0.40) \\
Individual & -0.30(0.62) & \bfseries -0.01(0.13) \\
\bottomrule
\end{tabular}

%% file: r-jnc-mf1.tex
\begin{tabular}{llS[table-format=1.2(1.2)]S[table-format=1.2(1.2)]}
\toprule
{} & {Test} & {Group} & {Individual} \\
{Model} & {Train} & {} & {} \\
\midrule
\multirow[c]{2}{*}{LCP-CWI} & Group & 0.47 & 0.47(0.09) \\
 & Individual & 0.46(0.17) & 0.47(0.12) \\
\hline
\multirow[c]{2}{*}{CWI} & Group & 0.73 & 0.63(0.06) \\
 & Individual & \bfseries 0.73(0.01) & \bfseries 0.64(0.06) \\
\bottomrule
\end{tabular}

%% file: r-tubelex-jc-r2.tex
\begin{tabular}{lS[table-format=-1.2(1.2)]S[table-format=-1.2(1.2)]}
\toprule
{Test} & {Group} & {Individual} \\
{Train} & {} & {} \\
\midrule
Group & \bfseries 0.41 & -0.10(0.41) \\
Individual & -0.06(0.64) & \bfseries 0.13(0.15) \\
\bottomrule
\end{tabular}

%% file: r-tubelex-jc-mf1.tex
\begin{tabular}{llS[table-format=1.2(1.2)]S[table-format=1.2(1.2)]}
\toprule
{} & {Test} & {Group} & {Individual} \\
{Model} & {Train} & {} & {} \\
\midrule
\multirow[c]{2}{*}{LCP-CWI} & Group & 0.71 & 0.65(0.05) \\
 & Individual & 0.56(0.18) & 0.56(0.11) \\
\hline
\multirow[c]{2}{*}{CWI} & Group & \bfseries 0.78 & 0.67(0.06) \\
 & Individual & 0.77(0.01) & \bfseries 0.67(0.04) \\
\bottomrule
\end{tabular}

%% file: r-jc-r2.tex
\begin{tabular}{lS[table-format=-1.2(1.2)]S[table-format=-1.2(1.2)]}
\toprule
{Test} & {Group} & {Individual} \\
{Train} & {} & {} \\
\midrule
Group & \bfseries -0.00 & -0.32(0.39) \\
Individual & -0.43(0.58) & \bfseries -0.09(0.13) \\
\bottomrule
\end{tabular}

%% file: r-jc-mf1.tex
\begin{tabular}{llS[table-format=1.2(1.2)]S[table-format=1.2(1.2)]}
\toprule
{} & {Test} & {Group} & {Individual} \\
{Model} & {Train} & {} & {} \\
\midrule
\multirow[c]{2}{*}{LCP-CWI} & Group & 0.34 & 0.35(0.08) \\
 & Individual & 0.34(0.01) & 0.37(0.07) \\
\hline
\multirow[c]{2}{*}{CWI} & Group & \bfseries 0.62 & 0.59(0.06) \\
 & Individual & 0.62(0.01) & \bfseries 0.59(0.07) \\
\bottomrule
\end{tabular}